\documentclass{article}

\PassOptionsToPackage{numbers, compress}{natbib}

\usepackage[preprint]{neurips_2023}

\usepackage{amsmath}
\usepackage{amssymb}
\usepackage{mathtools}
\usepackage{amsthm}

\usepackage{graphicx}
\usepackage{amsmath}
\usepackage{amssymb}
\usepackage{booktabs}
\usepackage{adjustbox}
\usepackage{boldline}
\usepackage{pifont}
\usepackage{xcolor}

\usepackage[utf8]{inputenc} %
\usepackage[T1]{fontenc}    %
\usepackage{hyperref}       %
\usepackage{url}            %
\usepackage{booktabs}       %
\usepackage{amsfonts}       %
\usepackage{nicefrac}       %
\usepackage{microtype}      %
\usepackage{xcolor}         %
\usepackage[capitalize,noabbrev]{cleveref}
\theoremstyle{plain}

\theoremstyle{definition}

\theoremstyle{remark}

\usepackage{wrapfig,lipsum,booktabs}

\newcommand{\red}[1]{{\color{red}{#1}}}
\newcommand{\green}[1]{{\color{green}{#1}}}

\newcommand{\cmark}{\green{\ding{51}}}%
\newcommand{\xmark}{\red{\ding{55}}}%

\usepackage{multirow}
\usepackage{slashbox}
\usepackage{diagbox}
\usepackage{dsfont}
\usepackage{algorithm}
\usepackage{algpseudocode}
\usepackage{algorithmicx}
\usepackage{enumitem}
\usepackage{caption}
\usepackage{subcaption}
\usepackage{abdalmageed_style}
\algrenewcommand\algorithmicrequire{\textbf{Input:}}
\algrenewcommand\algorithmicensure{\textbf{Output:}}

\newcommand{\rewrite}[1]{{\color{black}{#1}}}

\newcommand{\wrong}[1]{{\color{red}{#1}}}

\newcommand{\ind}{\perp\!\!\!\!\perp} 

\title{Shadow Datasets, New challenging datasets for Causal Representation Learning}

\author{
Jiageng Zhu$^{1,3}$ \qquad Hanchen Xie$^{2,3}$ \qquad Jianhua Wu$^2$ \qquad Jiazhi Li$^{1,3}$ \AND
  \qquad Mahyar Khayatkhoei$^{3}$ \qquad Mohamed E. Hussein $^{3}$ \qquad Wael AbdAlmageed$^{1,2,3}$  \\
$^1$ USC Ming Hsieh Department of Electrical and Computer Engineering \\
$^2$ USC Thomas Lord Department of Computer Science, Los Angeles USA\\
$^3$ USC Information Sciences Institute \\
\tt\small  \{jiagengz, hanchenx, wamageed\}@isi.edu}

\begin{document}

\maketitle

\begin{abstract}

Discovering causal relations among semantic factors is an emergent topic in representation learning.  Most causal representation learning (CRL) methods are fully supervised, which is impractical due to costly labeling.  To resolve this restriction, weakly supervised CRL methods were introduced. To evaluate CRL performance, four existing datasets, Pendulum, Flow, CelebA(BEARD) and CelebA(SMILE), are utilized.  However, existing CRL datasets are limited to simple graphs with few generative factors. Thus we propose two new datasets with a larger number of diverse generative factors and more sophisticated causal graphs. In addition, current real datasets, CelebA(BEARD) and CelebA(SMILE), the originally proposed causal graphs are not aligned with the dataset distributions. Thus, we propose modifications to them.
The Shadow datasets can be downloded through \url{https://github.com/Jiagengzhu/Shadow-dataset-for-crl}

\end{abstract}
\section{Introduction}
\label{sec:introduction}

Utilizing causality to address computer vision problems, such as distribution shifts \cite{bib:ood,sun2021recovering}, domain adaptation \cite{bib:domain-adaptation} and fairness \cite{Zhu2020Causal}, has become an emerging topic. To overcome this impediment, causal representation learning (CRL) \cite{bib:towards-causal,DBLP:conf/cvpr/YangLCSHW21,bib:do-VAE} has been introduced to simultaneously learn semantically meaningful representations, as well as identify the causal relations among these semantic representations, from high dimensional raw observations.
CausalVAE \cite{DBLP:conf/cvpr/YangLCSHW21}, for example,  utilizes a linear causal layer in a supervised learning fashion to discover causal structure. However, models like CausalVAE that need ground-truth labels during training has limited applicability becasue it is considerably expensive to annotate all factors of variations and their causal relationships in large scale datasets. To mitigate the limitations of fully supervised learning, DoVAE \cite{bib:do-VAE} introduced a weak supervision signal by applying the \emph{do-operation} \cite{reason:Pearl09a} to an input pair in the latent space to form reconstructions that  match the original inputs.

Current standard benchmarks are not sophisticated enough to comprehensively evaluate CRL methods because they are either extremely simple or improperly designed. Two synthetic datasets: Pendulum \cite{DBLP:conf/cvpr/YangLCSHW21} and Flow \cite{DBLP:conf/cvpr/YangLCSHW21}, and two real datasets: CelebA(SMILE) \cite{DBLP:conf/cvpr/YangLCSHW21} and CelebA(BEARD) \cite{DBLP:conf/cvpr/YangLCSHW21}, have been used in the literature \cite{DBLP:conf/cvpr/YangLCSHW21,bib:do-VAE}. However, these four benchmarks contain only a small number of generative factors, which can only construct simple causal graphs. Further, we observe that the originally proposed ground-truth causal graphs of CelebA(BEARD) and CelebA(SMILE) are not properly aligned with their statistical distributions.  Motivated by the limitations of existing benchmarks, we propose two novel datasets: \emph{Shadow-Sunlight} and \emph{Shadow-Pointlight}, which simulate the causal relations between (both sun and point-source) light, objects and shadows. Both datasets contain larger numbers of generative factors than existing datasets, which naturally leads to more complex causal graphs. Furthermore, to address the aforementioned flaw in CelebA(BEARD) and CelebA(SMILE), we also propose curating CelebA(BEARD) to align its statistical distribution with the ground-truth causal graph. CelebA(SMILE), on the other hand, assumes that Gender is a cause of Eyes open, which may be ethically concerning. Hence, we suggest avoiding CRL evaluations on this dataset.
To evaluate the performance of CRL methods, we utilize the the metrics proposed by \cite{bib:do-VAE} (i.e., PosMIC/TIC, and NegMIC/TIC) which measure the correctness/incorrectness of the causal relations.

\begin{figure*}
    \centering
     \includegraphics[width=0.72\textwidth]{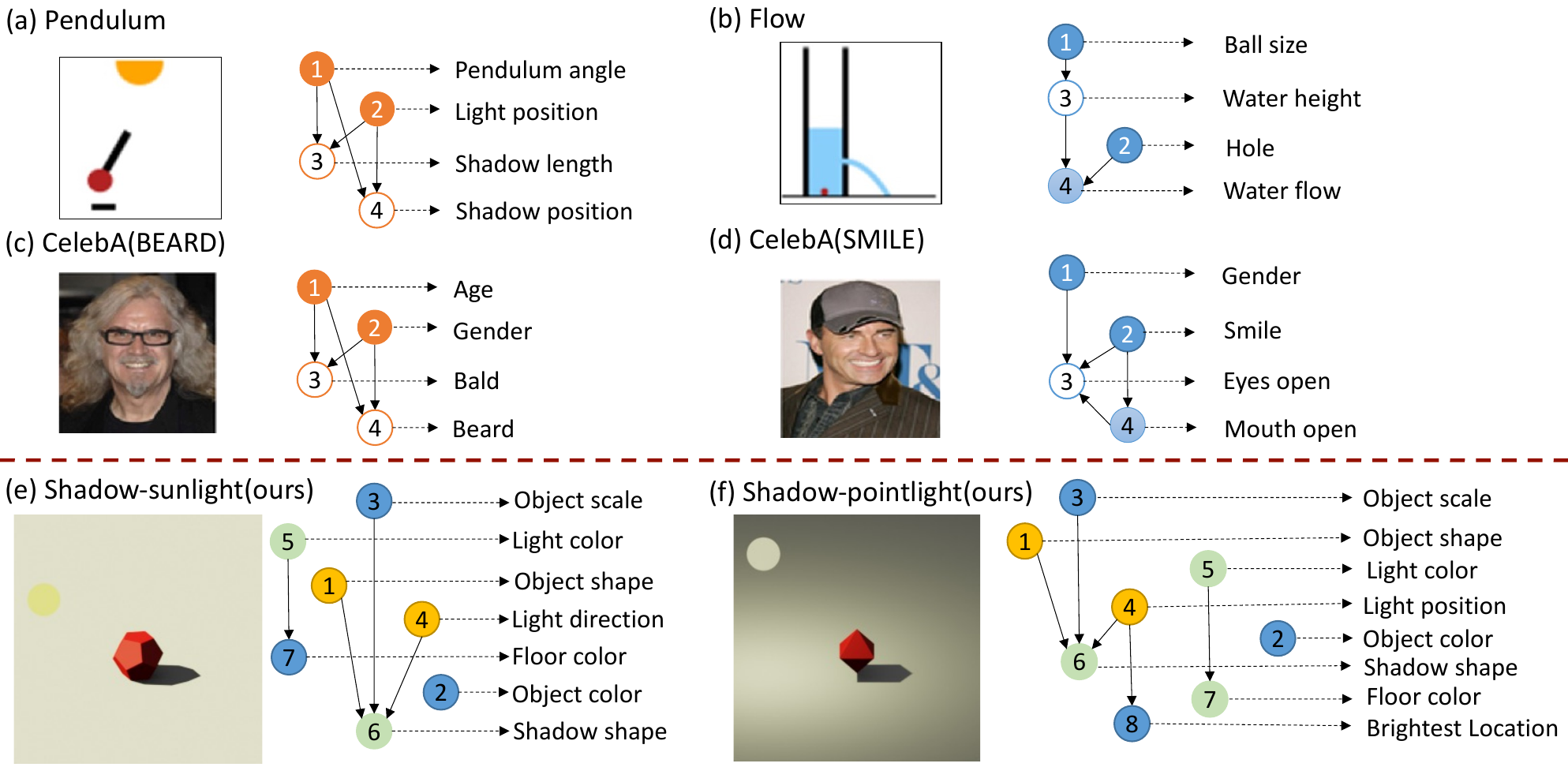}
    \caption{\textbf{Datasets:} Samples and ground-truth causal graphs of exisiting datasets and proposed new datasets. First two rows show existing datasets. In the third row, we provide samples and causal graphs of our newly proposed datasets: \emph{Shadow-Sunlight} and \emph{Shadow-Pointlight}.} %
    \label{fig:dataset}
\end{figure*}

The contributions of this paper are: 

\begin{itemize}[noitemsep,topsep=0pt,parsep=0pt,partopsep=0pt]
    \item \emph{Shadow CRL benchmarks}, which contain more generative factors than current CRL datasets and more sophisticated causal relations
    \item Identifying shortcomings of existing real-world datasets, and propose curating CelebA(BEARD), and suggest omitting CRL evaluations on CelebA(SMILE)
    \item Comprehensive evaluation on both existing datasets and proposed datasets to show the bigger challenge brought by new proposed datasets.
\end{itemize}

\section{Related work}
\noindent\textbf{Representation disentanglement and CRL:} \rewrite{Disentangled representation learning assumes that latent factors are mutually independent \cite{DBLP:conf/iclr/HigginsMPBGBML17}}, \rewrite{whereas CRL seeks to find the causal relations between factor pairs \cite{bib:towards-causal}.} 
Variational Autoencoder (VAE) \cite{DBLP:journals/corr/KingmaW13} has commonly been used for both tasks, which proposes disentangling the representation by minimizing reconstruction loss with Kullback-Leibler divergence ($D_{KL}$) as a regularizer on the latent space.
Despite previous attempts of seeking disentangled representation via unsupervised VAE-based models \cite{DBLP:conf/iclr/HigginsMPBGBML17,chen2019isolating,bib:laddervae}, supervision is proved to be necessary to identify the relations between latent factor pairs \cite{locatello2019challenging}.
\rewrite{To fulfill such requirement, CausalVAE \cite{DBLP:conf/cvpr/YangLCSHW21} directly uses the generative factor labels and adopts a linear causal discovery layer to discover the causal relations.} DoVAE \cite{bib:do-VAE} relaxes the need for strong supervision  by \rewrite{leveraging the \emph{do-operation} \cite{reason:Pearl09a}  to construct the latent representation space. }Brehmer et al. \cite{brehmer2022weakly} propose learning the causal representations by first learning disentangled exogenous factors, which requires both observational data and interventional data. However, generating perfectly interventional data requires full access to the underlying causal mechanism, which can not be guaranteed in real-world scenarios. Thus, we will not compare with \cite{brehmer2022weakly} because in this work, only observational data is taken as input.

\begin{table*}[]
\centering
\caption{Meta-data comparison between Shadow datsets and existing datasets}
\label{table:dataset-compare}
\renewcommand{\arraystretch}{1.2}
\begin{adjustbox}{width=0.75\textwidth}
\begin{tabular}{ccccccc}
\hlineB{3}
Dataset           & Resolution       & \begin{tabular}[c]{@{}c@{}}Factors of \\ Variation\end{tabular} & \begin{tabular}[c]{@{}c@{}}Number of \\ Samples\end{tabular} & Nuisance             & \begin{tabular}[c]{@{}c@{}}Various generative\\ factors \end{tabular} & \begin{tabular}[c]{@{}c@{}}Consistent with \\ Causal Graph\end{tabular} \\ \hline
Pendulum \cite{DBLP:conf/cvpr/YangLCSHW21}           & 96  $\times$ 96  & 4                                                               & 7000                                                    & \xmark & \cmark                                          & \cmark                                                   \\
Flow \cite{DBLP:conf/cvpr/YangLCSHW21}               & 96  $\times$ 96  & 4                                                               & 7000                                                    & \xmark & \cmark                                          & \cmark                                                   \\
CelebA(BEARD) \cite{DBLP:conf/cvpr/YangLCSHW21}     & 128 $\times$ 128 & 4                                                               & 202599                                                  & \cmark & \xmark                                          & \xmark                                                   \\
CelebA(SMILE) \cite{DBLP:conf/cvpr/YangLCSHW21}     & 128 $\times$ 128 & 4                                                               & 202599                                                  & \cmark & \xmark                                          & \xmark                                                   \\
\emph{Shadow-Sunlight}   & 128 $\times$ 128 & 7                                                               & 41160                                                   & \cmark & \cmark                                          & \cmark                                                   \\
\emph{Shadow-Pointlight} & 128 $\times$ 128 & 8                                                               & 41160                                                   & \cmark & \cmark                                          & \cmark                                                   \\ \hlineB{3}
\end{tabular}
\end{adjustbox}
\end{table*}
\noindent\textbf{CRL benchmarks:}
As shown in \cref{fig:dataset,table:dataset-compare}, the standard benchmarks for evaluating CRL methods are Pendulum, Flow, CelebA(BEARD) and CelebA(SMILE) \cite{DBLP:conf/cvpr/YangLCSHW21}.  Pendulum and Flow are synthetic datasets, while CelebA(BREAD) and CelebA(SMILE) are derivatives of CelebA dataset \cite{liu2015faceattributes}. Each of these datasets contains four causal generative factors. Pendulum focuses on pendulum angle, light position, shadow length, and shadow position. Flow focuses on ball size, water height, hole, and flow. CelebA(BEARD) focuses on age, gender, bald, and beard. CelebA(SMILE) focuses on gender, smile, eyes open, and mouth open. No nuisance factors exist in the two synthetic datasets. In the two real datasets, the number of nuisance factors (36) is much larger than the number of causal generative factors (4).
Furthermore, all generative factors in the real datasets are binary in nature.

\noindent\textbf{Structure learning:}
\rewrite{Structure learning aims to learn causal graph structure from census data, which} 
can roughly be sorted as constraint-based, score-based, and continuous optimization-based methods. Constraint-based methods \cite{spirtes2000constructing,kalisch2007estimating} explore the causal relations by testing conditional independence between factors. 
Score-based methods \cite{bib:BIC,bib:BGE,bib:MDL} use scoring functions to compare causal graph candidates.  Continuous optimization-based methods \cite{bib:notears,bib:dag-gnn,bib:gae} adapt deep neural network (DNN) with an optimization objective.
NOTEARS \cite{bib:notears} learns linear causal relationships through minimizing
the reconstruction loss with acyclicity constraint $h(A) = tr(e^{A\odot A})-d$ to enforce the causal graph, represented by the adjacency matrix $A$, to be directed acyclic, where $d$ is the total number of factors. GOLEM \cite{DBLP:conf/nips/NgG020} improves the causal discovery performance by proposing a new loss function that directly maximizes the empirical likelihood. 
DAG-GNN \cite{bib:dag-gnn} extends NOTEARS by introducing a graph neural network in order to estimate non-linear causal relations.  
Graph Autoencoders (GAE) \cite{bib:gae} were shown to significantly improve the performance of both DAG-GNN and NOTEARS.

\section{Shadow datasets }
\label{sec:datasets}

As illustrated in \cref{fig:dataset}, existing datasets, both synthetic and real, are too simple for CRL because they have few factors of variations, which cannot be used to construct a complex causal graph.
We propose two novel datasets: \emph{Shadow-Sunlight} and \emph{Shadow-Pointlight}, which contain a larger number of  factors of variation, which naturally leads to more complex causal graphs, as shown in \cref{sec:shadow}.

\subsection{Shadow benchmarks}
\label{sec:shadow}
Existing datasets are too simple  because they only contain four generative factors, which cannot construct sophisticated causal relations. To address this limitation, inspired by the Pendulum \cite{DBLP:conf/cvpr/YangLCSHW21}, we propose two novel datasets: \emph{Shadow-Sunlight} and \emph{Shadow-Pointlight}, which contain seven and eight factors of variation, respectively, so that more complex causal graphs can be constructed.
Further, compared with the binary generative factors in CelebA(SMILE) and CelebA(BREAD), the generative factors in Shadow datasets are more diverse. 
A detailed comparison is shown in   \cref{table:dataset-compare}.

Shown in \cref{fig:dataset}, Shadow datasets are generated using Blender \cite{bib:blender} with the Cycle rendering engine. The proposed datasets simulate the causal relations between light, object, floor, and shadow. In \emph{Shadow-Sunlight}, we set the light source type to be \emph{sunlight}, which emits parallel light rays. In \emph{Shadow-Pointlight}, we set the light source type to be \emph{point light}, in which all light rays are emitted from a single point. Due to the different attributes of the two types of light sources, the causal mechanisms inherent in the two environments are different, which leads to two distinct datasets.  \rewrite{The object attributes in the Shadow datasets are the cross product of seven different object shapes, seven different object colors, and seven different scales.} Further, there are six different light colors. In \emph{Shadow-Sunlight}, there are 20 different light directions; and in \emph{Shadow-Pointlight}, the light directions are controlled by 20 different light positions. Since the shadow shape, the floor color, and the brightest floor position are effect factors, their values are controlled by object shape, object scale, light position/direction, and light color.
Object color serves as the nuisance factor, which has no causal relation with other factors. Compared with Shadow datasets, Pendulum and Flow datasets do not contain any nuisance factors. Contrarily, the number of nuisance factors of the real datasets (36) is much larger than the number of causal generative factors (4), which necessitates using ground-truth labels to focus on the expected factors during training \cite{bib:dual-disentangle}.

\subsection{Generation of Shadow datasets using Blender}
\label{sec:dataset-detail}
\begin{figure}[htb]
    \centering
     \includegraphics[width=0.5\textwidth]{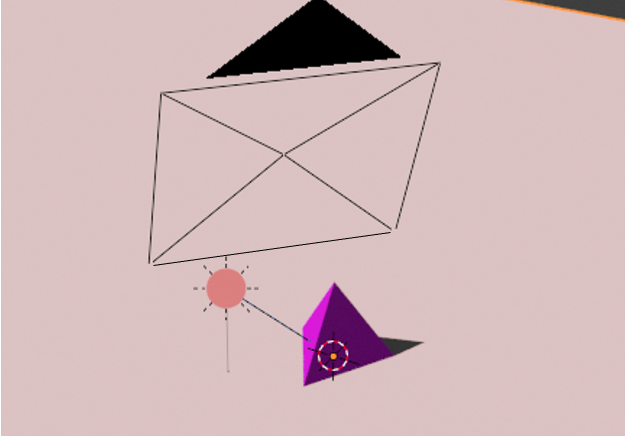}
    \caption{Environment scene in Blender.} %
    \label{fig:blender-overview}
\end{figure}

\begin{figure}[]
    \centering
     \includegraphics[width=0.45\textwidth]{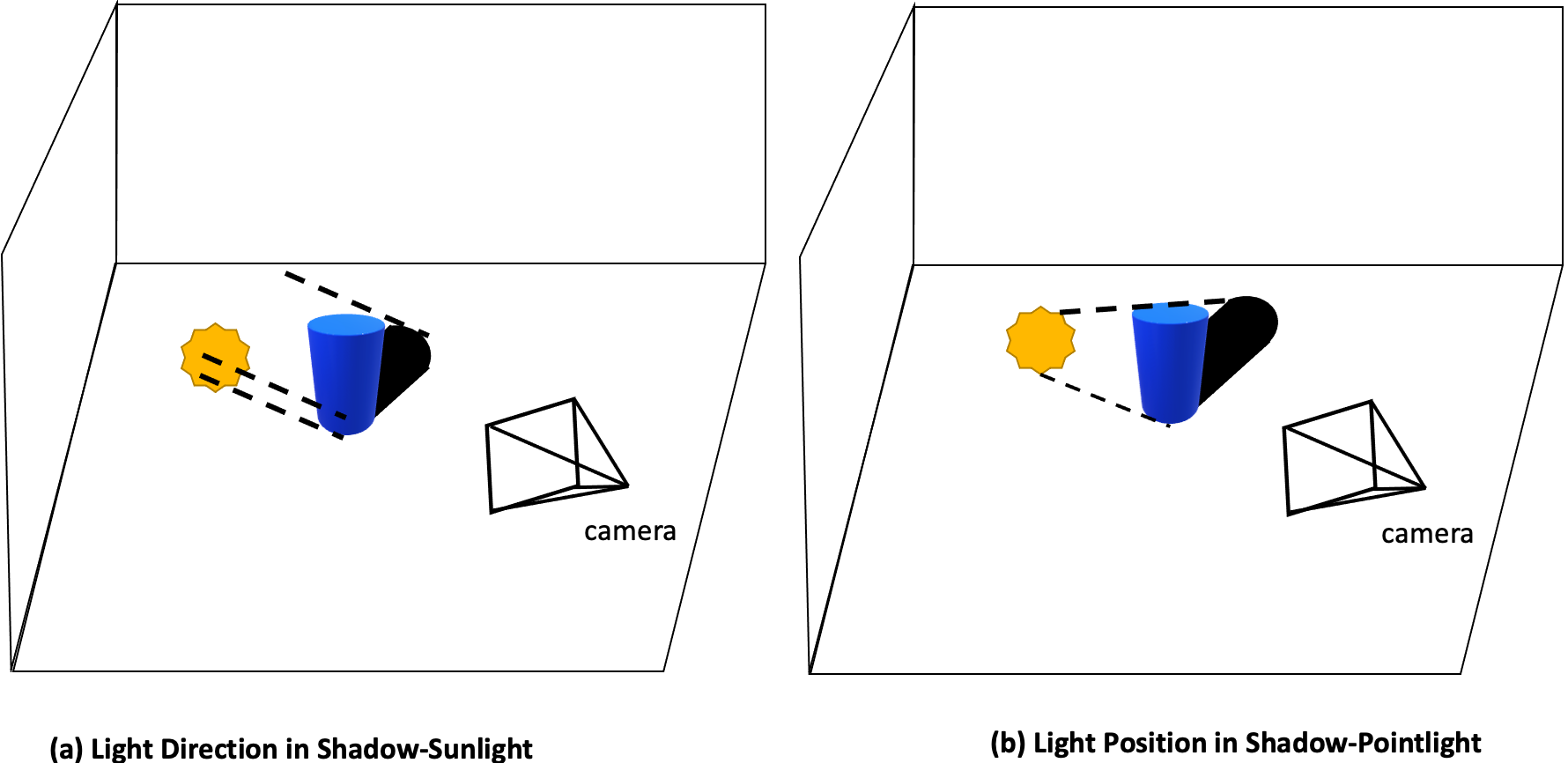}
    \caption{Illustrations of two different light sources in Shadow-Sunlight and Shadow-Pointlight, respectively.} %
    \label{fig:light-sources}
\end{figure}

In this section, we describe the scene setting in Blender and how we set the generative values for each factor. The overview of the scene in Blender is shown in
\cref{fig:blender-overview}, where the black box ($\boxtimes$) is the camera. For Shadow-Pointlight, the light position is value determined by the center of the light ball. For Shadow-Sunlight, since all rays are parallel, we set the direction from the center of light ball to the origin of scene to be the value of sunlight direction, the difference of two types of light sources in Shadow-Sunlight and Shadow-pointlight can be illustrated in \cref{fig:light-sources}. For the scale, we set the basic size (Object scale=1), which is determined by the height from the center of object gravity to the floor. Other object scales are determined by changing this height.

\subsection{Factors of Variant}
\begin{figure*}
    \centering
     \includegraphics[width=0.87\textwidth]{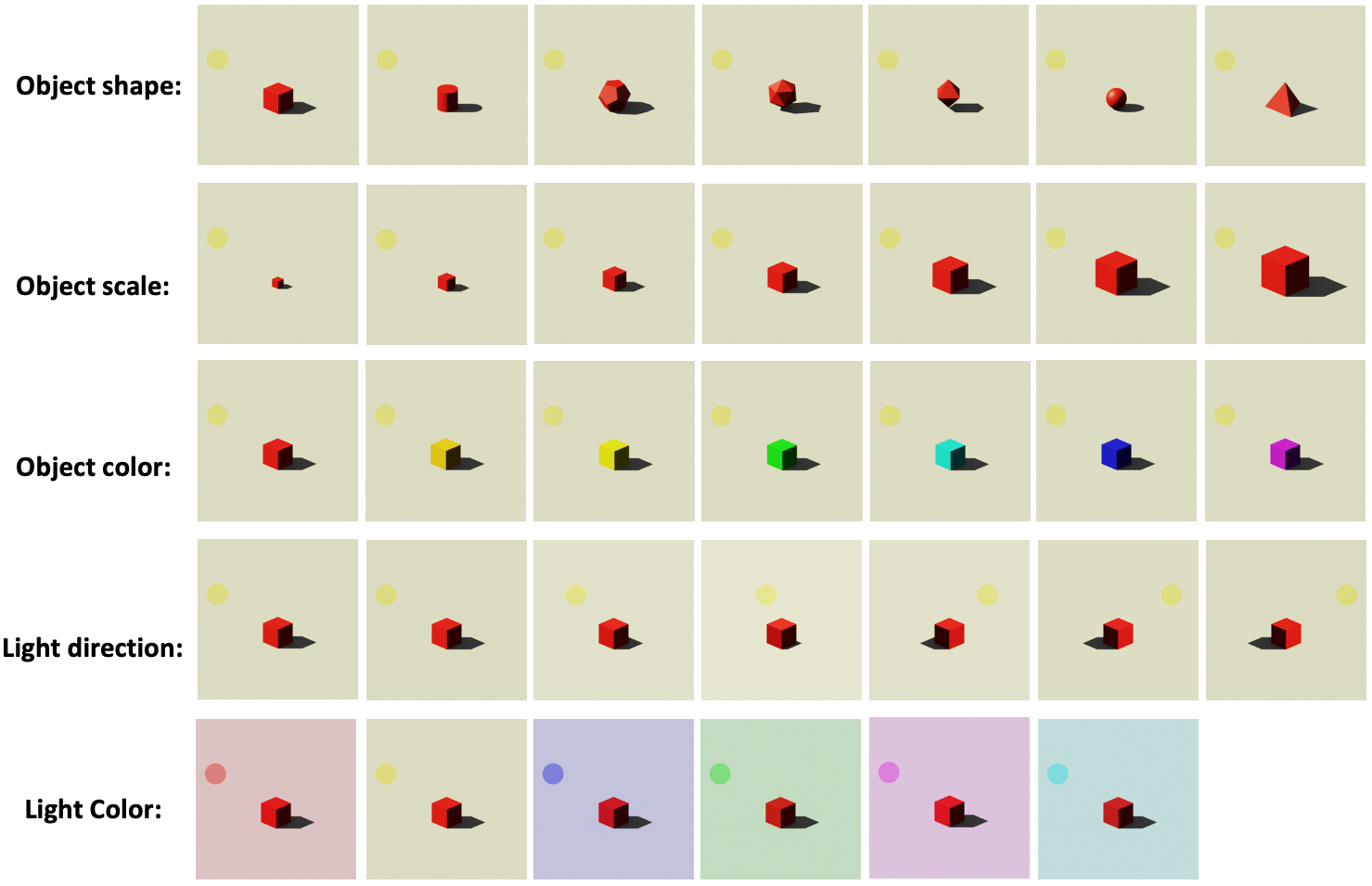}
    \caption{Factors of variation in Shadow-Sunlight. There are 20 different values of Light direction factors.} %
    \label{fig:sunlight-generative}
\end{figure*}

\begin{figure*}
    \centering
     \includegraphics[width=0.87\textwidth]{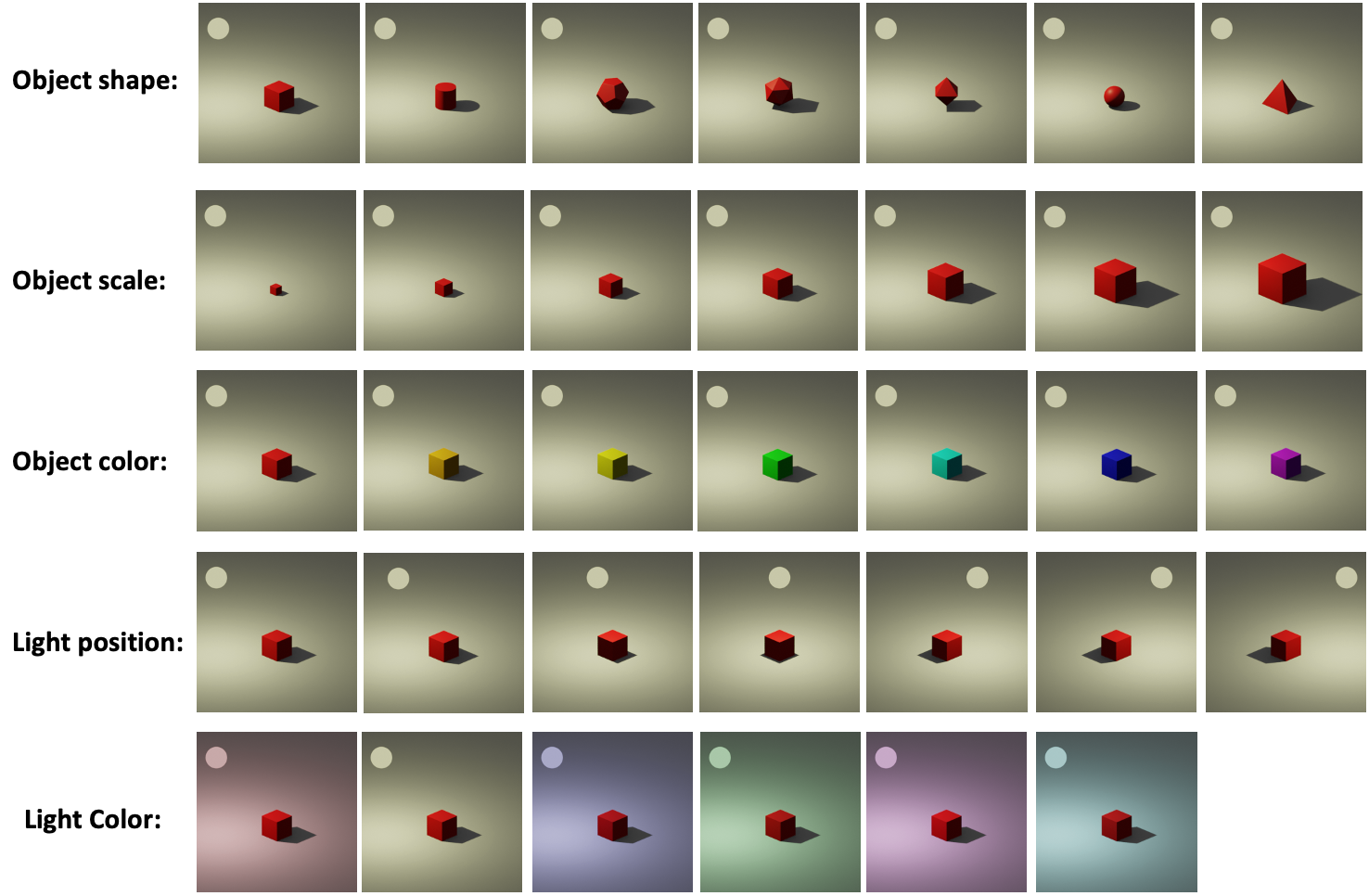}
    \caption{Factors of variation in Shadow-Pointlight. There are 20 different values of Light position factors.} %
    \label{fig:pointlight-generative}
\end{figure*}

As introduced in \cref{sec:datasets}, Shadow-Sunlight and Shadow-Pointlight have seven factors of variant and eight factors of variant, respectively. The name of each factor of variant and the number of each variant are summarized in \cref{table:settings}. Shadow datasets contain seven different object shapes: Cube, Sphere, Cylinder, Tetrahedron,140
Octahedron, Dodecahedron, and Icosahedron. Objects can have one of seven different colors in141
(R,G,B): Red (255,0,0), Orange (255,128,0), Yellow (255,255,0), Green (0,255,0), Cyan (0,255,255),142
Blue (0,0,255), and Purple (128,0,255). Each object can be assigned one of seven different scales.143
There are six different light colors (R,B,G): SkyBlue (128,255,255), Plum (255,128,255), Khaki144
(255,255,128), Lavender (128,128,255), Lightgreen (128,255,128), and Coral (255,128,128). In145
Shadow-Sunlight, there are 20 different light directions and in Shadow-Pointlight, the light directions146
are controlled by 20 different light positions. Since the shadow shape, the floor color, and the147
brightest floor position are effect factors, their values are controlled by object shape, object scale,148
light position/direction, and light color. To better illustrate each generative factor of variation, we show the traversal of each generative factor of Shadow-Sunlight in \cref{fig:sunlight-generative} and each generative factor of Shadow-Pointlight in \cref{fig:pointlight-generative}.

\begin{table}[htb]
\centering
\caption{Generative factor settings of Shadow-Sunlight and Shadow-Pointlight. Shadow shape, Floor color and Brightest floor position are effect factors controlled by cause factors.}
\label{table:settings}
\renewcommand{\arraystretch}{1.2}
\begin{adjustbox}{width=0.5\textwidth}
\begin{tabular}{c|c}
\hlineB{2}
Factors                                                                                 & Number of Variants \\ \hline
Object shape                                                                            & 7                  \\ \hline
Object color                                                                            & 7                  \\ \hline
Object scale                                                                            & 7                  \\ \hline
Light direction / position                                                              & 20                 \\ \hline
Light color                                                                             & 6                  \\ \hline
Shadow shape                                                                            &  N/A            \\ \hline
Floor color                                                                             &   N/A             \\ \hline
\begin{tabular}[c]{@{}c@{}}Brightest floor position\\ (only in pointlight) \end{tabular} &    N/A              \\ \hlineB{2}
\end{tabular}
\end{adjustbox}
\end{table}


\subsection{Split of training and test}
For both Shadow-Sunlight and Shadow-Pointlight, we split the dataset by taking $90\%$ samples for training, which are $36160$ samples for training. The remaining $10\%$ samples are used for testing, which are $4116$ testing samples.

\section{Modifications to real-world benchmarks}
\label{sec:modification}

CelebA(BEARD) and CelebA(SMILE) are two real-world datasets for CRL \cite{DBLP:conf/cvpr/YangLCSHW21}. The originally proposed causal graphs are shown in \cref{fig:dataset}(c) and (d). We examine the statistical relations between factor pairs in these datasets and discovered inconsistencies between these relations and the originally proposed causal graphs. Thus, learning the optimal causal graphs from the original data is ill-posed.

As illustrated in \cite{reason:Pearl09a}, conditional independence tests can assess the correctness of a causal graph. Thus,  we utilize the $\chi^2$ test to evaluate the correctness of CelebA(BEARD)'s originally proposed causal graph \cite{DBLP:conf/cvpr/YangLCSHW21}, which assumes that Age and Gender are mutually independent. Besides, since Bald and Beard are \emph{colliders} of Age and Gender, Age and Gender should be dependent when conditioned on Bald or Beard. Further, Bald and Beard should be mutually dependent. By performing conditional independence tests on the original CelebA(BEARD), we find that Age and Gender are not mutually independent because the $p$-value for $\chi^2$ test is $10^{-5}$, which is much smaller than the  significance level $\alpha=0.01$. In a similar way, we find that other conditional independence relations are valid. Thus, we randomly remove samples to make Gender and Age statistically independent and create a curated CelebA(BEARD) to align the originally proposed causal graph. More details are in \cref{sec:dataset-detail}. \rewrite{For the CelebA(SMILE), we omit the evaluation on it and encourage the research community to do the same because of its ethically troubling assumption that gender is a cause of eyes to be opened or not. }

\subsection{Conditionally Independent Test on Original Real Datasets and  Curate CelebA(BEARD)}
\begin{figure*}[htb]
    \centering
    \includegraphics[width=0.94\textwidth]{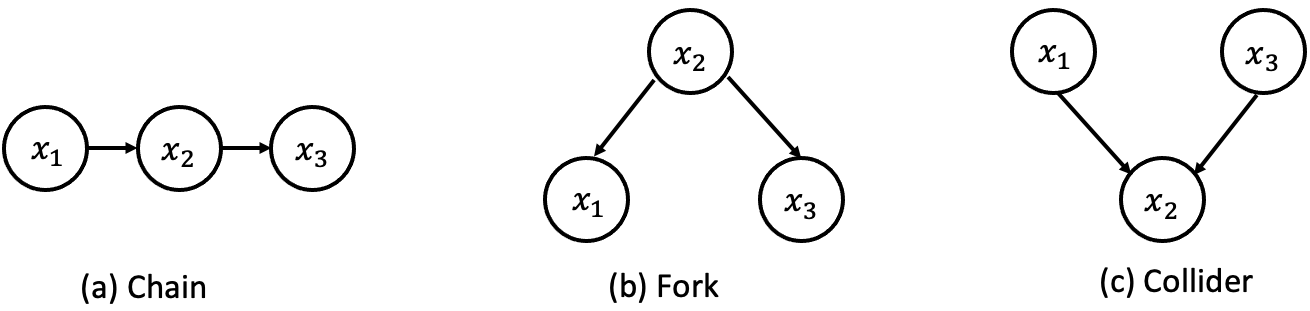} 
    \caption{Three types of causal graph building blocks.}
    \label{fig:graph-blocks}
\end{figure*}

As discussed in \cref{sec:modification} and proposed by \cite{reason:Pearl09a}, the correctness of a proposed causal graph can be justified by testing the conditional independence between two factors. In causality theory, There are three types of graph building blocks: \emph{chain}, \emph{fork} and \emph{immortality (collider)}. The structures of them are illustrated in \Cref{fig:graph-blocks}.

As shown in \Cref{fig:graph-blocks}(a) and (b) respectively, \emph{Chain}  and \emph{fork} share the same set of dependencies. In both structures, $X_1$ and $X_3$ are associated, i.e., they are mutually dependent with each other. If we condition on $X_2$, $(X_1|X_2)$ and $(X_3|X_2)$ become conditional independent with each other, i.e., $(X_1|X_2) \ind (X_3|X_2)$.

Different from \emph{chain} and \emph{fork},  \emph{Collider (immorality)} has distinct set of dependencies, where the structure of \emph{collider} is shown in \cref{fig:graph-blocks}(c). $X_1$ and $X_3$ are independent with each other but will become conditional dependent if both of them are conditioned on $X_2$, i.e., $(X_1|X_2) \not\!\perp\!\!\!\perp (X_3|X_2)$. Besides, when conditioned on descendants of $X_2$, 
$(X_1|de(X_2)) $ and $ (X_3|de(X_2))$ are also conditionally dependent, i.e., $(X_1|de(X_2)) \not\!\perp\!\!\!\perp (X_3|de(X_2))$, where $de(X_2)$ stands for descendants of $X_2$.

By utilizing the property of each basic block, we can determine the conditional independent relations between each factor according to the original proposed causal graphs of CelebA(BEARD) and CelebA(SMILE)  \cite{DBLP:conf/cvpr/YangLCSHW21}, which are shown in \cref{fig:dataset}(c) and (d). Although we discussed in \cref{sec:modification} that we suggest omitting the evaluation on CelebA(SMILE), we also test the conditionally independent relations in CelebA(SMILE) and observe that data distribution  of CelebA(SMILE) is also not consistent with the originally proposed causal graph.
Discussed in \cref{sec:modification}, we incorporate $\chi^2$ test to assess the conditional independent relations, where we set the significant level to be $\alpha = 0.05$. The null hypothesis is set as $H_0$: two factors are independent. Contrarily, $H_1$ is: two factors are not independent. If the $p$-value less than the significant level $\alpha=0.05$, we reject $H_0$.
Since all generative factors in CelebA are binary, the degree of freedom is $1$.
We describe the test for CelebA(BEARD) and CelebA(SMILE) test in following paragraphs sequentially.
\begin{table}[htb]
\centering
\caption{$p$-value of $\chi^2$ (freedom=1) test between each factor when no condition on original CelebA(BEARD). The \wrong{red number} indicates inconsistent relations between two factors with original proposed causal graph.} 
\label{table:chi-original-celebA-no-condition}
\begin{tabular}{|c|c|c|c|c|}
\hline
       & Age  & Gender & Bald & Beard \\ \hline
Age    &      & \wrong{ $10^{-5}$}   & 0    & 0     \\ \hline
Gender & \wrong{ $10^{-5}$} &        & 0    & 0     \\ \hline
Bald   & 0    & 0      &      & 0     \\ \hline
Beard  & 0    & 0      & 0    &       \\ \hline
\end{tabular}
\end{table}
\begin{table}[]
\centering
\caption{$p$-value of $\chi^2$ (freedom=1) test between Gender and Age when conditioned on different factors on original CelebA(BEARD).} 
\label{table:chi-original-celebA-beard-condition}
\begin{tabular}{|c|c|}
\hline
Conditioned on   & p-value \\ \hline
Bald           & 0.0019  \\ \hline
Beard          & 0.0     \\ \hline
Bald and Beard & 0.1519  \\ \hline
\end{tabular}
\end{table}

\noindent\textbf{CelebA(BEARD) conditionally independent tests}
The originally proposed causal graph of CelebA(BEARD) is shown in \cref{fig:dataset}(c), where Bald and Beard are colliders of Age and Gender. By utilizing the conditionally independent relations inherent in three basic blocks in causal graph, according to the originally proposed causal graph of CelebA(BEARD), the conditionally independent relations are summarised as: (1) Age and Gender should be mutually independent with each other, (2) Age and Gender become conditionally dependent when conditioned on Bald or Beard, (3) Bald and Beard are mutually dependent with each other, (4) Bald and Beard are conditionally independent  when conditioned on both Age and Beard, (5) Any other remaining relations are dependent relations. Results of $\chi^2$ tests between two factors when they are not conditioned on any other factors are shown in \cref{table:chi-original-celebA-no-condition}. Since the $p$-value between Age and Gender is smaller than significant level $\alpha=0.05$, which indicates that the Age and Gender are not mutually independent with each other. 
As shown in \cref{table:chi-original-celebA-beard-condition}, other conditionally independent tests demonstrate the consistency between collider structure of original proposed causal graph with data distribution.

\begin{table}[]
\centering
\caption{$p$-value of $\chi^2$ (freedom=1) test between each factor when no condition on original CelebA(SMILE). The \wrong{red number} indicates inconsistent relations between two factors with original proposed causal graph.} 
\label{table:chi-original-celebA-smile-no-condition}
\begin{adjustbox}{width=0.45\textwidth}
\begin{tabular}{|c|c|c|c|c|}
\hline
           & Gender & Smile & Eyes open & Mouth open \\ \hline
Gender     &        & \wrong{  $3 \times 10^{-5}$}  & 0.02      & 0          \\ \hline
Smile      & \wrong{  $3 \times 10^{-5}$}   &       & 0         & 0          \\ \hline
Eyes open  & 0.02   & 0     &           & 0          \\ \hline
Mouth open & 0      & 0     & 0         &            \\ \hline
\end{tabular}
\end{adjustbox}
\end{table}
\begin{table}[]
\centering
\caption{$p$-value of $\chi^2$ (freedom=1) test between Gender and Mouth open when conditioned on different factors on original CelebA(SMILE).} 
\label{table:chi-original-celebA-smile-condition}

\begin{tabular}{|c|c|}
\hline
Conditioned on & p-value            \\ \hline
Eyes open    & 0.0017             \\ \hline
Smile        & \wrong{$7 \times 10^{-6}$} \\ \hline
\end{tabular}
\end{table}
\noindent\textbf{CelebA(SMILE) conditionally independent tests}
Similar with the test in CelebA(SMILE), according to the originally proposed causal graph of CelebA(SMILE), which is shown in \cref{fig:dataset}(d), the corresponding conditionally independent relations are: (1) Gender and Smile are mutually independent, (2) Gender and Mouth open are mutually independent ,(3) Gender and Smile are conditionally dependent when conditioned on Eyes open, (4) Gender and Mouth open are conditionally dependent when conditioned on Eyes open, (5) any other remaining relations are dependent relations.  Results of $\chi^2$ tests between two factors when there is no condition are shown in \cref{table:chi-original-celebA-smile-no-condition}. When conditioned on Smile, according to the original proposed CelebA(SMILE) causal graph, the Gender and Mouth open should be independent, which is not aligned with the data distribution shown in \cref{table:chi-original-celebA-smile-condition}.

\begin{table}[]
\centering
\caption{ $p$-value of $\chi^2$ (freedom=1) test between each factor when no condition on curated CelebA(BEARD).} 
\label{table:chi-curated-celebA-no-condition}
\begin{tabular}{|c|c|c|c|c|}
\hline
       & Age & Gender & Bald & Beard \\ \hline
Age    &     & 0.5    & 0    & 0     \\ \hline
Gender & 0.5 &        & 0    & 0     \\ \hline
Bald   & 0   & 0      &      & 0     \\ \hline
Beard  & 0   & 0      & 0    &       \\ \hline
\end{tabular}
\end{table}
\begin{table}[]
\centering
\caption{$p$-value of $\chi^2$ (freedom=1) test between Gender and Age when conditioned on different factors on curate CelebA(BEARD).} 
\label{table:chi-curated-celebA-beard-condition}
\begin{tabular}{|c|c|}
\hline
Conditioned on   & p-value \\ \hline
Bald           & 0.0  \\ \hline
Beard          & 0.0     \\ \hline
Bald and Beard & 0.2459  \\ \hline
\end{tabular}
\end{table}
\noindent\textbf{Curate CelebA(BEARD) and  conditionally independent tests on curate CelebA(BEARD)}
As shown by the conditionally independent tests about the original CelebA(BEARD) dataset, the only inconsistent relation is the relation between Gender and Age, where they are supposed to be mutually independent according to the causal graph. Thus, to address this issue, we explicitly sample the uniform Gender, where a random sample is first chosen, then according to the Gender and Age of that sample, we sample another random sample with the same Age but opposite gender. By performing this curate, in the curate CelebA(BEARD), we can make Gender and Age to be statistically independent with each other, which is consistent with the originally proposed causal graph.

By applying this curate operation, we explicitly make the Gender and Age to be mutually independent with each other.
To justify the consistency between the data distribution in curate CelebA(BEARD) and the originally proposed causal graph, we conduct the same conditionally independent tests on curated CelebA(BEARD). As shown in \cref{table:chi-curated-celebA-no-condition} and \cref{table:chi-curated-celebA-beard-condition}, the data distributions in curated CelebA(BEARD) are aligned with originally proposed causal graph. Thus, the curated CelebA(BEARD) can be used to evaluate causal representation learning.

\section{Experimental evaluation}
\subsection{Benchmarks}
\label{sec:benchmark}
\noindent\textbf{Datasets:}
To empirically demonstrate the performance of our method, we conduct experiments on the Shadow datasets, discussed in \cref{sec:datasets}. Furthermore, we show that our method outperforms SOTA methods, i.e., CausalVAE \cite{DBLP:conf/cvpr/YangLCSHW21} and DoVAE \cite{bib:do-VAE}, on the standard Pendulum and Flow \cite{DBLP:conf/cvpr/YangLCSHW21} benchmarks. For evaluation on real-world datasets, we include experimental results on the modified CelebA(BEARD) as discussed in \cref{sec:modification}. In CelebA(BEARD), since the number of nuisance factors (36) is much larger than the number of factors for causal inference (4), the ground-truth labels are required to focus on the expected attributes \cite{bib:dual-disentangle}. Thus, experiments are conducted in a semi-supervised setting, where the methods are trained with $\{10\%, 20\%, 30\%, 40\% \}$ of labeled data.

\begin{table*}[]
\centering
\caption{Causal representation metrics tested on Pendulum and Flow.  }
\label{table:synthetic}
\renewcommand{\arraystretch}{1.3}
{
\begin{adjustbox}{width=0.95\textwidth}
\setlength{\tabcolsep}{0.3em}
\begin{tabular}{ccccccccccccc}
\hlineB{3}
\multicolumn{1}{c|}{\multirow{2}{*}{Models}} & \multicolumn{6}{c}{Pendulum}                                                                                                                                            & \multicolumn{6}{c}{Flow}                                                                                                                                           \\ \cline{2-13} 
\multicolumn{1}{c|}{}                        & PosMIC $\uparrow$                & PosTIC $\uparrow$                & NegMIC $\downarrow$                 & NegTIC  $\downarrow$               &  $F_1^{MIC}$ $\uparrow$                 & \multicolumn{1}{c|}{ $F_1^{TIC}$ $\uparrow$}                  & PosMIC $\uparrow$                & PosTIC  $\uparrow$               & NegMIC  $\downarrow$                                & NegTIC  $\downarrow$               &  $F_1^{MIC}$ $\uparrow$                 &  $F_1^{TIC}$ $\uparrow$                 \\ \hline
\multicolumn{13}{c}{Fully Supervised learning methods (all labels are used)}                                                                                                                                                                                                                                                                                                                \\ \hline
\multicolumn{1}{c|}{CausalVAE \cite{DBLP:conf/cvpr/YangLCSHW21}}               & 53.0$\pm$4.5           & 43.4$\pm$3.7           & 46.6$\pm$3.9           & 37.0$\pm$4.2          & 53.2$\pm$3.6           & \multicolumn{1}{c|}{51.4$\pm$3.2}           & 45.1 $\pm$4.8          & 36.7 $\pm$4.2          & 43.3 $\pm$5.1                          & 33.7 $\pm$3.2         & 50.2 $\pm$4.4          & 47.3 $\pm$3.7          \\
\multicolumn{1}{c|}{ConditionVAE \cite{bib:conditionvae}}            & 36.5$\pm$3.0           & 27.8$\pm$3.2           & 34.6$\pm$4.2           & 25.7 $\pm$3.6         & 46.9 $\pm$4.7          & \multicolumn{1}{c|}{40.5 $\pm$3.5}          & 28.6 $\pm$3.2          & 21.3 $\pm$3.1          & 27.2 $\pm$2.8                          & 20.6 $\pm$2.7         & 41.1 $\pm$5.1          & 33.6 $\pm$4.0          \\ \hline
\multicolumn{13}{c}{Unsupervised  Learning methods (no label is used)}                                                                                                                                                                                                                                                                                                                      \\ \hline
\multicolumn{1}{c|}{CausalVAE(unsup) \cite{DBLP:conf/cvpr/YangLCSHW21}}        & 20.5 $\pm$2.6          & 11.8 $\pm$2.7          & 23.3 $\pm$3.2          & 14.7 $\pm$1.9         & 32.4 $\pm$3.4          & \multicolumn{1}{c|}{20.7 $\pm$3.1}          & 22.8 $\pm$2.7          & 12.5 $\pm$1.4          & 21.5 $\pm$2.4                          & 12.0 $\pm$1.9         & 35.3 $\pm$5.6          & 21.9 $\pm$4.7          \\
\multicolumn{1}{c|}{$\beta$-VAE \cite{DBLP:conf/iclr/HigginsMPBGBML17}}                 & 21.2 $\pm$2.7          & 12.7 $\pm$2.9          & 23.7 $\pm$3.1          & 12.6 $\pm$1.9         & 33.2 $\pm$3.3          & \multicolumn{1}{c|}{22.2 $\pm$2.7}          & 23.6 $\pm$3.6          & 12.5 $\pm$1.9          & 22.1 $\pm$2.5                          & 11.4 $\pm$1.9         & 36.2 $\pm$4.9          & 21.9 $\pm$4.2          \\
\multicolumn{1}{c|}{LadderVAE \cite{bib:laddervae}}               & 15.2 $\pm$1.9          & 8.6 $\pm$1.0           & \textbf{14.2 $\pm$1.7} & \textbf{7.9 $\pm$0.9} & 25.8 $\pm$3.0          & \multicolumn{1}{c|}{15.7 $\pm$2.8}          & 16.2 $\pm$1.8          & 10.5 $\pm$1.0          & \textbf{13.3 $\pm$1.2} & \textbf{6.9 $\pm$0.6} & 27.3 $\pm$ 3.2         & 18.9 $\pm$2.8          \\ \hline
\multicolumn{13}{c}{Reduced supervision method (no label is used; supervision source is image pairing )}                                                                                                                                                                                                                                                                                    \\ \hline
\multicolumn{1}{c|}{Do-VAE \cite{bib:do-VAE}}                  & 54.1 $\pm$4.5          & 44.0 $\pm$4.2          & 40.2 $\pm$3.9          & 31.6 $\pm$3.2         & 56.8 $\pm$5.2          & \multicolumn{1}{c|}{53.6 $\pm$4.3}          & 50.7 $\pm$4.7          & 41.3 $\pm$4.2          & 36.8 $\pm$3.8                          & 27.2 $\pm$3.0         & 56.3 $\pm$5.9          & 52.7 $\pm$4.9

 \\ \hlineB{3}
\end{tabular}

\end{adjustbox}
}
\vspace{-\baselineskip}
\end{table*}

\noindent\textbf{Metrics:}
As illustrated in \cite{bib:do-VAE}, mutual information coefficient (MIC) and total information coefficient (TIC), used in CausalVAE \cite{DBLP:conf/cvpr/YangLCSHW21}, have fundamental limitations, and PosMIC/TIC and NegMIC/TIC \cite{bib:do-VAE} can better reveal the correctness of the causal structure, which leads to a more proper assessment of CRL.

\begin{figure*}
    \centering
     \includegraphics[width=0.85\textwidth]{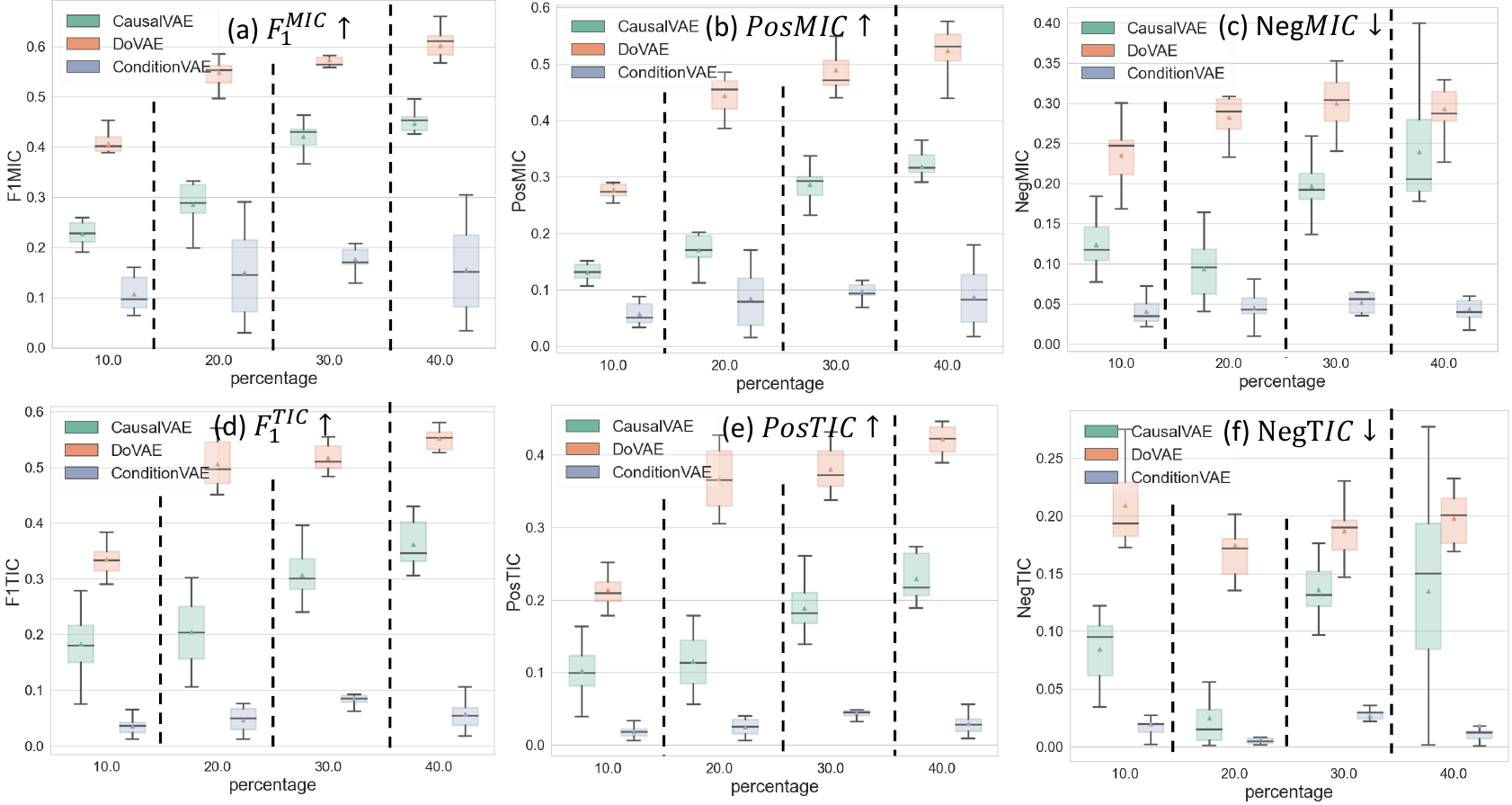}
    \caption{Tested on the curated CelebA(BEARD), our method consistently outperforms SOTAs.} %
    \label{fig:box-plot-celebA-beard}
\end{figure*}
To calculate \emph{positive} metrics, latent effect factors before the causal discovery layer, which is GAE in our method, are first set to zero. Then, MIC and TIC between the latent effect factors after the causal discovery layer and the ground-truth generative effect factors are calculated to be the final score of PosMIC and PosTIC. Higher metrics scores indicate better performance because an effect value is expected to be correctly inferred from its causes.
In contrast, NegMIC and NegTIC assess the falseness of causal effects by firstly setting the latent cause factors before causal discovery layer to zero, then 
the MIC and TIC between the latent cause factors after the causal discovery layer and the ground-truth generative cause factors is calculated to be the final score. Since causal relations should only propagate from causes to effects \cite{reason:Pearl09a}, the lower scores of Neg metrics indicate better performance.
Further, to combine both \emph{positive} and \emph{negative} metrics,  $F_1^{MIC}$ and $F_1^{TIC}$ \cite{bib:do-VAE} is used to calculate the harmonic mean between the positive metrics and one minus the negative metrics. 
All metrics range from zero to one, and we scale all scores by 100 in our tables for demonstration.
\begin{table*}[]
\centering
\caption{Causal representation metrics tested on Shadow-Sunlight and Shadow-Pointlight.  }
\label{table:shadows}
\renewcommand{\arraystretch}{1.3}
{
\begin{adjustbox}{width=0.95\textwidth}
\setlength{\tabcolsep}{0.3em}
\begin{tabular}{ccccccccccccc}
\hlineB{3}
\multicolumn{1}{c|}{\multirow{2}{*}{Models}} & \multicolumn{6}{c|}{Shadow-Sunlight}                                                                                                                                                                                                                       & \multicolumn{6}{c}{Shadow-Pointlight}                                                                                                                                                                                                                     \\ \cline{2-13} 
\multicolumn{1}{c|}{}                        & PosMIC $\uparrow$                                    & PosTIC $\uparrow$                                    & NegMIC  $\downarrow$                          & NegTIC $\downarrow$                            &  $F_1^{MIC}$ $\uparrow$                                     & \multicolumn{1}{c|}{ $F_1^{TIC}$ $\uparrow$}                  & PosMIC $\uparrow$                                    & PosTIC  $\uparrow$                                   & NegMIC  $\downarrow$                           & NegTIC   $\downarrow$                          &  $F_1^{MIC}$ $\uparrow$                                     &  $F_1^{TIC}$ $\uparrow$                                     \\ \hline
\multicolumn{13}{c}{Fully Supervised learning methods (all labels are used)}                                                                                                                                                                                                                                                                                                                                                                                                                                                                                          \\ \hline
\multicolumn{1}{c|}{CausalVAE \cite{DBLP:conf/cvpr/YangLCSHW21}}               & 33.6$\pm$5.2                               & 25.6$\pm$3.8                               & 43.1$\pm$6.9                      & 32.5$\pm$4.5                      & 39.1$\pm$5.6                               & \multicolumn{1}{c|}{34.9$\pm$4.2}           & 39.2 $\pm$3.7                              & 29.1 $\pm$4.5                              & 45.1 $\pm$5.1                     & 35.1 $\pm$3.8                     & 44.2 $\pm$4.6                              & 39.2 $\pm$4.7                              \\
\multicolumn{1}{c|}{ConditionVAE \cite{bib:conditionvae}}            & 18.5$\pm$3.3                               & 10.6$\pm$2.8                               & 25.8$\pm$4.1                      & 15.7 $\pm$5.1                     & 29.4 $\pm$4.1                              & \multicolumn{1}{c|}{18.6$\pm$3.1}           & 11.6 $\pm$2.5                              & 4.8 $\pm$2.2                               & 14.4 $\pm$2.6                     & 6.6 $\pm$2.2                      & 20.3 $\pm$3.8                              & 12.7 $\pm$3.1                              \\ \hline
\multicolumn{13}{c}{Unsupervised  Learning methods (no label is used)}                                                                                                                                                                                                                                                                                                                                                                                                                                                                                                \\ \hline
\multicolumn{1}{c|}{CausalVAE(unsup) \cite{DBLP:conf/cvpr/YangLCSHW21}}        & 13.2 $\pm$3.3                              & 7.6 $\pm$2.3                               & 17.5$\pm$3.2                      & 9.7 $\pm$3.9                      & 22.4 $\pm$4.7                              & \multicolumn{1}{c|}{14.4 $\pm$4.8}          & 12.5 $\pm$2.8                              & 5.6 $\pm$1.5                               & 14.9 $\pm$2.4                     & 7.3 $\pm$2.0                      & 19.7 $\pm$3.6                              & 10.9 $\pm$3.5                              \\
\multicolumn{1}{c|}{$\beta$-VAE \cite{DBLP:conf/iclr/HigginsMPBGBML17}}                 & 12.7 $\pm$4.6                              & 6.7 $\pm$4.2                               & 18.8 $\pm$4.3                     & 11.2 $\pm$3.7                     & 21.6 $\pm$3.6                              & \multicolumn{1}{c|}{12.2 $\pm$3.3}          & 11.3 $\pm$3.3                              & 4.8 $\pm$1.6                               & \textbf{14.6 $\pm$2.5}            & \textbf{7.1 $\pm$2.2}             & 19.8 $\pm$5.0                              & 11.7 $\pm$3.8                              \\
\multicolumn{1}{c|}{LadderVAE \cite{bib:laddervae}}               & 13.7 $\pm$3.1                              & 6.0 $\pm$3.5                               & \textbf{17.2 $\pm$4.7}            & \textbf{10.7 $\pm$ 2.9}           & 22.8 $\pm$4.5                              & \multicolumn{1}{c|}{11.1 $\pm$5.1}          & 7.9 $\pm$4.1                               & 5.2 $\pm$1.9                               & 15.2 $\pm$3.2                     & 8.7 $\pm$2.6                      & 14.3 $\pm$ 2.6                             & 9.8 $\pm$2.1                               \\ \hline
\multicolumn{13}{c}{Reduced supervision method (no label is used; supervision source is image pairing )}                                                                                                                                                                                                                                                                                                                                                                                                                                                              \\ \hline
\multicolumn{1}{c|}{DoVAE \cite{bib:do-VAE}}                   & 32.6 $\pm$3.5                              & 26.7 $\pm$4.1                              & 31.8 $\pm$2.5                     & 24.0 $\pm$2.8                     & 43.9 $\pm$3.3                              & \multicolumn{1}{c|}{34.8 $\pm$3.8}          & 34.7 $\pm$4.4                              & 30.8 $\pm$4.3                              & 40.1 $\pm$4.8                     & 29.7 $\pm$3.3                     & 43.3 $\pm$4.4                              & 40.1 $\pm$3.8                              \\

 \hlineB{3}
\end{tabular}

\end{adjustbox}
}
\vspace{-\baselineskip}
\end{table*}

\subsection{SOTAs method comparision}
Comparisons between the proposed and SOTA methods are shown in \cref{table:synthetic,table:shadows} and \cref{fig:box-plot-celebA-beard}. 
CausalVAE \cite{DBLP:conf/cvpr/YangLCSHW21} and ConditionVAE \cite{bib:conditionvae} are both fully supervised methods, which are limited by the necessity of ground-truth labels. CausalVAE limits the causal discovery layer to be linear, thus showing sub-optimal performance. ConditionVAE forces latent factors to be mutually independent so that, although ConditionVAE achieves good results on NegMIC and NegTIC, it cannot discover the correct causal relations. Thus, it shows unsatisfactory performance on PosMIC and PosTIC, which jeopardizes $F_1^{MIC}$ and $F_1^{TIC}$. For unsupervised methods, CausalVAE(unsup) \cite{DBLP:conf/cvpr/YangLCSHW21}, $\beta$-VAE \cite{DBLP:conf/iclr/HigginsMPBGBML17} and LadderVAE \cite{bib:laddervae}, in spite of their good results on NegMIC and NegTIC, they show unsatisfactory performances on PosMIC, PosTIC, $F_1^{MIC}$, and $F_1^{TIC}$ because they cannot learn the expected latent representation and encode semantic information \cite{locatello2019challenging}.  DoVAE \cite{bib:do-VAE}, which also depends on weak supervision, merely relies on an underdeveloped GAE for discovering causal relations during training so that it may fail when a large number of causal factors exists, as shown in \cref{table:shadows}. As shown in \cref{table:synthetic,table:shadows}, compared with existing datasets, Shadow datasets bring larger challenges for current CRL methods, where both correct causal relations become hard to find and false relations become hard to prevent.

\section{Conclusion}
In this work, to bring harder challenges to CRL, we propose Shadow datasets, which simulate the causal relations between shadow and objects that are illuminated by different light source types.   Further, we observe issues of the existing real-world datasets and propose a fix by curating them. Thorough studies show the potential benefit of the proposed datasets for future research in this field.

\clearpage
{\small
\bibliographystyle{icml2023}
\bibliography{egbib}

\begin{thebibliography}{27}
\providecommand{\natexlab}[1]{#1}
\providecommand{\url}[1]{\texttt{#1}}
\expandafter\ifx\csname urlstyle\endcsname\relax
  \providecommand{\doi}[1]{doi: #1}\else
  \providecommand{\doi}{doi: \begingroup \urlstyle{rm}\Url}\fi

\bibitem[Brehmer et~al.(2022)Brehmer, De~Haan, Lippe, and
  Cohen]{brehmer2022weakly}
Brehmer, J., De~Haan, P., Lippe, P., and Cohen, T.
\newblock Weakly supervised causal representation learning.
\newblock In \emph{Advances in Neural Information Processing Systems}, 2022.

\bibitem[Chen et~al.(2019)Chen, Li, Grosse, and Duvenaud]{chen2019isolating}
Chen, R. T.~Q., Li, X., Grosse, R., and Duvenaud, D.
\newblock Isolating sources of disentanglement in variational autoencoders,
  2019.

\bibitem[Community(2022)]{bib:blender}
Community, B.~O.
\newblock \emph{Blender - a 3D modelling and rendering package}.
\newblock Blender Foundation, Stichting Blender Foundation, Amsterdam, 2022.
\newblock URL \url{https://www.blender.org}.

\bibitem[Feng et~al.(2018)Feng, Wang, Ke, Zeng, Tao, and
  Song]{bib:dual-disentangle}
Feng, Z., Wang, X., Ke, C., Zeng, A.-X., Tao, D., and Song, M.
\newblock Dual swap disentangling.
\newblock In Bengio, S., Wallach, H., Larochelle, H., Grauman, K.,
  Cesa-Bianchi, N., and Garnett, R. (eds.), \emph{Advances in Neural
  Information Processing Systems}, volume~31. Curran Associates, Inc., 2018.
\newblock URL
  \url{https://proceedings.neurips.cc/paper/2018/file/fdf1bc5669e8ff5ba45d02fded729feb-Paper.pdf}.

\bibitem[Geiger \& Heckerman(1994)Geiger and Heckerman]{bib:BIC}
Geiger, D. and Heckerman, D.
\newblock Learning gaussian networks.
\newblock In \emph{Proceedings of the Tenth International Conference on
  Uncertainty in Artificial Intelligence}, UAI'94, pp.\  235–243, San
  Francisco, CA, USA, 1994. Morgan Kaufmann Publishers Inc.
\newblock ISBN 1558603328.

\bibitem[Grünwald \& Vitányi(2008)Grünwald and Vitányi]{bib:MDL}
Grünwald, P.~D. and Vitányi, P.~M.
\newblock Algorithmic information theory.
\newblock In Adriaans, P. and {van Benthem}, J. (eds.), \emph{Philosophy of
  Information}, Handbook of the Philosophy of Science, pp.\  281--317.
  North-Holland, Amsterdam, 2008.
\newblock \doi{https://doi.org/10.1016/B978-0-444-51726-5.50013-3}.
\newblock URL
  \url{https://www.sciencedirect.com/science/article/pii/B9780444517265500133}.

\bibitem[Heckerman et~al.(1995)Heckerman, Geiger, and Chickering]{bib:BGE}
Heckerman, D., Geiger, D., and Chickering, D.~M.
\newblock Learning bayesian networks: The combination of knowledge and
  statistical data.
\newblock \emph{Machine Learning}, 20\penalty0 (3):\penalty0 197--243, Sep
  1995.
\newblock ISSN 1573-0565.
\newblock \doi{10.1023/A:1022623210503}.
\newblock URL \url{https://doi.org/10.1023/A:1022623210503}.

\bibitem[Higgins et~al.(2017)Higgins, Matthey, Pal, Burgess, Glorot, Botvinick,
  Mohamed, and Lerchner]{DBLP:conf/iclr/HigginsMPBGBML17}
Higgins, I., Matthey, L., Pal, A., Burgess, C., Glorot, X., Botvinick, M.,
  Mohamed, S., and Lerchner, A.
\newblock beta-vae: Learning basic visual concepts with a constrained
  variational framework.
\newblock In \emph{5th International Conference on Learning Representations,
  {ICLR} 2017, Toulon, France, April 24-26, 2017, Conference Track
  Proceedings}. OpenReview.net, 2017.
\newblock URL \url{https://openreview.net/forum?id=Sy2fzU9gl}.

\bibitem[Kalisch \& B{\"u}hlman(2007)Kalisch and
  B{\"u}hlman]{kalisch2007estimating}
Kalisch, M. and B{\"u}hlman, P.
\newblock Estimating high-dimensional directed acyclic graphs with the
  pc-algorithm.
\newblock \emph{Journal of Machine Learning Research}, 8\penalty0 (3), 2007.

\bibitem[Kingma \& Welling(2014)Kingma and
  Welling]{DBLP:journals/corr/KingmaW13}
Kingma, D.~P. and Welling, M.
\newblock Auto-encoding variational bayes.
\newblock In Bengio, Y. and LeCun, Y. (eds.), \emph{2nd International
  Conference on Learning Representations, {ICLR} 2014, Banff, AB, Canada, April
  14-16, 2014, Conference Track Proceedings}, 2014.
\newblock URL \url{http://arxiv.org/abs/1312.6114}.

\bibitem[Lee et~al.(2016)Lee, Sugiyama, von Luxburg, Guyon, and
  Garnett]{bib:laddervae}
Lee, D.~D., Sugiyama, M., von Luxburg, U., Guyon, I., and Garnett, R. (eds.).
\newblock \emph{Advances in Neural Information Processing Systems 29: Annual
  Conference on Neural Information Processing Systems 2016, December 5-10,
  2016, Barcelona, Spain}, 2016.
\newblock URL \url{https://proceedings.neurips.cc/paper/2016}.

\bibitem[Liu et~al.(2021)Liu, Sun, Wang, Tang, Li, Qin, Chen, and Liu]{bib:ood}
Liu, C., Sun, X., Wang, J., Tang, H., Li, T., Qin, T., Chen, W., and Liu, T.-Y.
\newblock Learning causal semantic representation for out-of-distribution
  prediction.
\newblock In Ranzato, M., Beygelzimer, A., Dauphin, Y., Liang, P., and Vaughan,
  J.~W. (eds.), \emph{Advances in Neural Information Processing Systems},
  volume~34, pp.\  6155--6170. Curran Associates, Inc., 2021.
\newblock URL
  \url{https://proceedings.neurips.cc/paper/2021/file/310614fca8fb8e5491295336298c340f-Paper.pdf}.

\bibitem[Liu et~al.(2015)Liu, Luo, Wang, and Tang]{liu2015faceattributes}
Liu, Z., Luo, P., Wang, X., and Tang, X.
\newblock Deep learning face attributes in the wild.
\newblock In \emph{Proceedings of International Conference on Computer Vision
  (ICCV)}, December 2015.

\bibitem[Locatello et~al.(2019)Locatello, Bauer, Lucic, Rätsch, Gelly,
  Schölkopf, and Bachem]{locatello2019challenging}
Locatello, F., Bauer, S., Lucic, M., Rätsch, G., Gelly, S., Schölkopf, B.,
  and Bachem, O.
\newblock Challenging common assumptions in the unsupervised learning of
  disentangled representations, 2019.

\bibitem[Ng et~al.(2019)Ng, Zhu, Chen, and Fang]{bib:gae}
Ng, I., Zhu, S., Chen, Z., and Fang, Z.
\newblock A graph autoencoder approach to causal structure learning.
\newblock \emph{arXiv preprint arXiv:1911.07420}, 2019.

\bibitem[Ng et~al.(2020)Ng, Ghassami, and Zhang]{DBLP:conf/nips/NgG020}
Ng, I., Ghassami, A., and Zhang, K.
\newblock On the role of sparsity and dag constraints for learning linear dags.
\newblock In \emph{NeurIPS}, 2020.
\newblock URL
  \url{https://proceedings.neurips.cc/paper/2020/hash/d04d42cdf14579cd294e5079e0745411-Abstract.html}.

\bibitem[Pearl(2009)]{reason:Pearl09a}
Pearl, J.
\newblock \emph{Causality: Models, Reasoning and Inference}.
\newblock Cambridge University Press, 2nd edition, 2009.

\bibitem[Schölkopf et~al.(2021)Schölkopf, Locatello, Bauer, Ke, Kalchbrenner,
  Goyal, and Bengio]{bib:towards-causal}
Schölkopf, B., Locatello, F., Bauer, S., Ke, N.~R., Kalchbrenner, N., Goyal,
  A., and Bengio, Y.
\newblock Toward causal representation learning.
\newblock \emph{Proceedings of the IEEE}, 109\penalty0 (5):\penalty0 612--634,
  2021.
\newblock \doi{10.1109/JPROC.2021.3058954}.

\bibitem[Sohn et~al.(2015)Sohn, Lee, and Yan]{bib:conditionvae}
Sohn, K., Lee, H., and Yan, X.
\newblock Learning structured output representation using deep conditional
  generative models.
\newblock In Cortes, C., Lawrence, N., Lee, D., Sugiyama, M., and Garnett, R.
  (eds.), \emph{Advances in Neural Information Processing Systems}, volume~28.
  Curran Associates, Inc., 2015.
\newblock URL
  \url{https://proceedings.neurips.cc/paper/2015/file/8d55a249e6baa5c06772297520da2051-Paper.pdf}.

\bibitem[Spirtes et~al.(2000)Spirtes, Glymour, Scheines, Kauffman, Aimale, and
  Wimberly]{spirtes2000constructing}
Spirtes, P., Glymour, C., Scheines, R., Kauffman, S., Aimale, V., and Wimberly,
  F.
\newblock Constructing bayesian network models of gene expression networks from
  microarray data.
\newblock 2000.

\bibitem[Sun et~al.(2021{\natexlab{a}})Sun, Wu, Zheng, Liu, Chen, Qin, and
  Liu]{bib:domain-adaptation}
Sun, X., Wu, B., Zheng, X., Liu, C., Chen, W., Qin, T., and Liu, T.-Y.
\newblock Recovering latent causal factor for generalization to distributional
  shifts.
\newblock In Ranzato, M., Beygelzimer, A., Dauphin, Y., Liang, P., and Vaughan,
  J.~W. (eds.), \emph{Advances in Neural Information Processing Systems},
  volume~34, pp.\  16846--16859. Curran Associates, Inc., 2021{\natexlab{a}}.
\newblock URL
  \url{https://proceedings.neurips.cc/paper/2021/file/8c6744c9d42ec2cb9e8885b54ff744d0-Paper.pdf}.

\bibitem[Sun et~al.(2021{\natexlab{b}})Sun, Wu, Zheng, Liu, Chen, Qin, and
  Liu]{sun2021recovering}
Sun, X., Wu, B., Zheng, X., Liu, C., Chen, W., Qin, T., and Liu, T.-Y.
\newblock Recovering latent causal factor for generalization to distributional
  shifts.
\newblock In \emph{NeurIPS 2021}, December 2021{\natexlab{b}}.

\bibitem[Yang et~al.(2021)Yang, Liu, Chen, Shen, Hao, and
  Wang]{DBLP:conf/cvpr/YangLCSHW21}
Yang, M., Liu, F., Chen, Z., Shen, X., Hao, J., and Wang, J.
\newblock Causalvae: Disentangled representation learning via neural structural
  causal models.
\newblock In \emph{{IEEE} Conference on Computer Vision and Pattern
  Recognition, {CVPR} 2021, virtual, June 19-25, 2021}, pp.\  9593--9602.
  Computer Vision Foundation / {IEEE}, 2021.
\newblock URL
  \url{https://openaccess.thecvf.com/content/CVPR2021/html/Yang\_CausalVAE\_Disentangled\_Representation\_Learning\_via\_Neural\_Structural\_Causal\_Models\_CVPR\_2021\_paper.html}.

\bibitem[Yu et~al.(2019)Yu, Chen, Gao, and Yu]{bib:dag-gnn}
Yu, Y., Chen, J., Gao, T., and Yu, M.
\newblock {DAG}-{GNN}: {DAG} structure learning with graph neural networks.
\newblock In Chaudhuri, K. and Salakhutdinov, R. (eds.), \emph{Proceedings of
  the 36th International Conference on Machine Learning}, volume~97 of
  \emph{Proceedings of Machine Learning Research}, pp.\  7154--7163. PMLR,
  09--15 Jun 2019.
\newblock URL \url{https://proceedings.mlr.press/v97/yu19a.html}.

\bibitem[Zheng et~al.(2018)Zheng, Aragam, Ravikumar, and Xing]{bib:notears}
Zheng, X., Aragam, B., Ravikumar, P., and Xing, E.~P.
\newblock {DAGs with NO TEARS: Continuous Optimization for Structure Learning}.
\newblock In \emph{Advances in Neural Information Processing Systems}, 2018.

\bibitem[Zhu et~al.(2022)Zhu, Xie, and AbdAlmgaeed]{bib:do-VAE}
Zhu, J., Xie, H., and AbdAlmgaeed, W.
\newblock Do-operation guided causal representation learning with reduced
  supervision strength.
\newblock In \emph{NeurIPS 2022 Workshop on Causality for Real-world Impact},
  2022.
\newblock URL \url{https://openreview.net/forum?id=KbjUEXmlKWJ}.

\bibitem[Zhu et~al.(2020)Zhu, Ng, and Chen]{Zhu2020Causal}
Zhu, S., Ng, I., and Chen, Z.
\newblock Causal discovery with reinforcement learning.
\newblock In \emph{International Conference on Learning Representations}, 2020.
\newblock URL \url{https://openreview.net/forum?id=S1g2skStPB}.

\end{thebibliography}
}

\end{document}